\documentclass[11pt,a4paper]{article}
\usepackage[hyperref]{emnlp2020}
\usepackage{times}
\usepackage{latexsym}

\usepackage{graphicx}
\usepackage{multirow}

\usepackage{floatrow}
\usepackage[font=small,labelfont=bf,tableposition=top]{caption}
\newfloatcommand{capbtabbox}{table}[][\FBwidth]

\usepackage{microtype}

\aclfinalcopy 

\title{Annotated Job Ads with Named Entity Recognition}

\newcommand*\samethanks[1][\value{footnote}]{\footnotemark[#1]}

\author{ 
    Felix Stollenwerk\thanks{~~Work done while working at Arbetsförmedlingen}~~\thanks{~~Correspondence to \texttt{felix.stollenwerk@ai.se}}
	\\
	AI Sweden\\
	\And
	Niklas Fastlund\\
	Arbetsförmedlingen\\
	\And
	Anna Nyqvist\samethanks[1] 
     \\
	ICA Gruppen\\
	\And
	Joey Öhman\samethanks[1]
	\\
	AI Sweden\\	
}

\date{}

\begin{document}
\maketitle

\begin{abstract}
We have trained a named entity recognition (NER) model that screens Swedish job ads for different kinds of useful information (e.g.  skills required from a job seeker).  It was obtained by fine-tuning KB-BERT. The biggest challenge we faced was the creation of a labelled dataset, which required manual annotation. This paper gives an overview of the methods we employed to make the annotation process more efficient and to ensure high quality data. We also report on the performance of the resulting model. 
\end{abstract}

\section{Introduction \label{sec:introduction}}

\begin{figure}[ht]
  \includegraphics[scale=0.42]{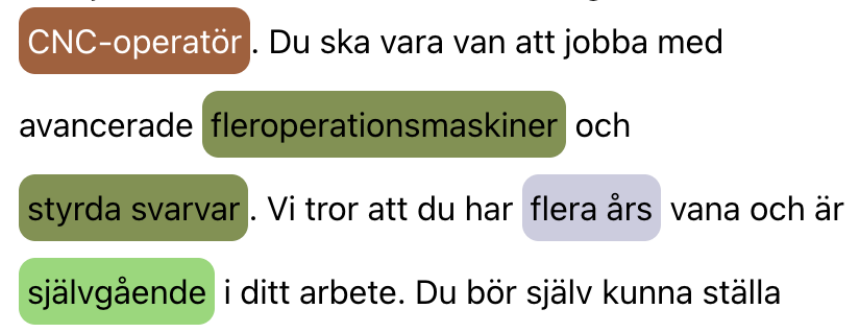}
  \caption{  
  Example of an annotated job ad. 
  The classes used in the excerpt are called \texttt{JOB\_TITLE} (brown), \texttt{SKILL\_HARD} (dark green), \texttt{SKILL\_SOFT} (light green) and \texttt{EXPERIENCE\_TIME} (light violet).}
  \label{fig:annotation_difficulty_examples}
\end{figure}

Natural language processing (NLP) has been subject to vast improvements in recent years.  Transformer-based models like BERT \citep{devlin2019bert} have pushed the boundaries of performance,  leveraging transfer learning in NLP through unsupervised pre-training of large language models and subsequent fine-tuning for a specific downstream task.
Swedish models are readily available \citep{swedish-bert} and have great potential to revolutionize the way text data is handled in the public and private sector in Sweden. 
However, despite those impressive leaps forwards, in order to create fine-tuned models that can be employed for real-world use cases,  labelled datasets are still required. These datasets usually need to be manually created and this often needs to be done in-house, for reasons such as that the annotation requires domain-expert knowledge or the data contains sensitive information.  
%
In our case, the goal was to create a dataset of annotated job ads in order to train a NER model which allows one to analyze job ads on a large scale. We aimed to detect a wide range of interesting information, initially represented by 16 classes. An example excerpt of an annotated job ad is shown in Fig.~\ref{fig:annotation_difficulty_examples}.

\begin{figure*}[ht]
    \centering
    \includegraphics[scale=0.32]{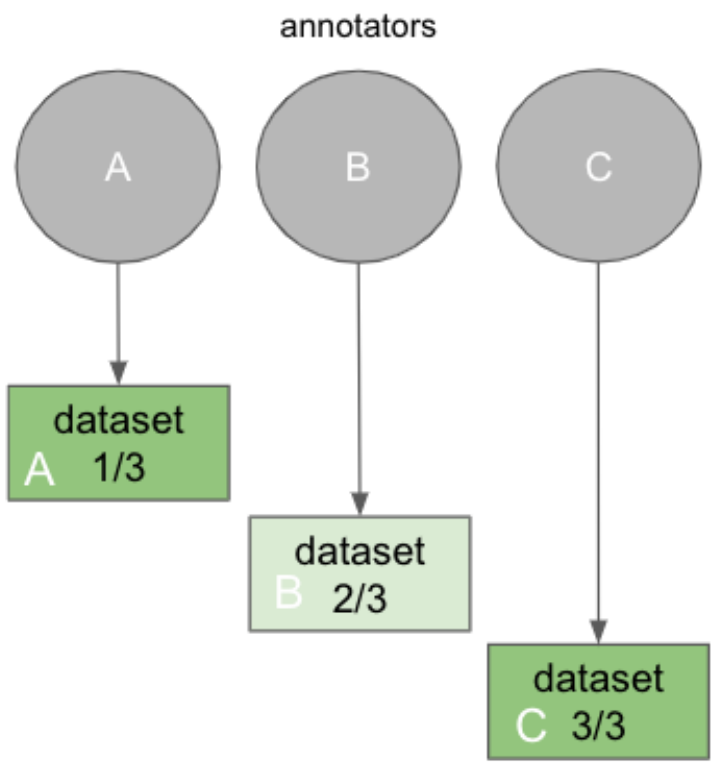} \hspace{2cm}
    \includegraphics[scale=0.29]{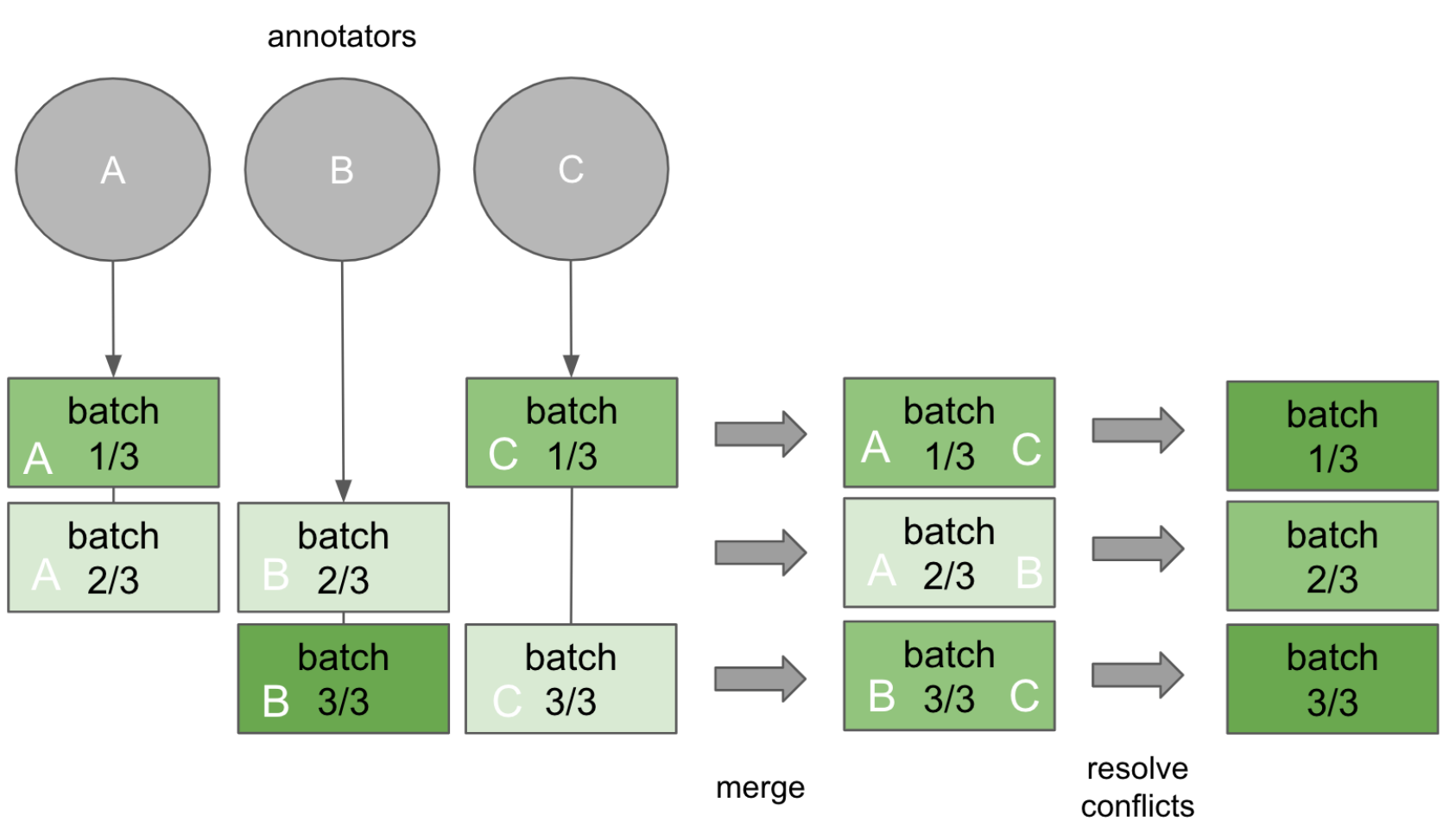} 
    \caption{\textit{Left:} Simple annotation strategy.  The whole dataset is divided into N parts for N annotators.  Each annotator takes care of their part independently, without any double-checking. The color of the parts represents the quality of the annotations (dark = high quality,  light = low quality) and the figure shows an example.  \textit{Right:} Our (more sophisticated) annotation strategy.  Firstly, each sample is annotated by two different annotators. After merging, the conflicts are discussed and resolved in a collective effort. A concrete example is shown in Fig.~\ref{fig:annotation_strategy_merge}.}
    \label{fig:annotation_strategies}     
\end{figure*}

The difficulty of the manual annotation process depends strongly on the use case and the complexity of the dataset one intends to create.  
A simple annotation task can often be handled straight forwardly. In the simplest possible set-up, the dataset is evenly split up between annotators who simply go through their respective shares and annotate each sample exactly once, see the left panel of Fig.~\ref{fig:annotation_strategies}.
However, in situations where the annotation task is more challenging, a simple strategy may be insufficient. There are mainly two sorts of potential problems:
\begin{enumerate}
 \item[A.] The effort to reach a sufficient amount of data (usually such that a model performance plateau is approached) exceeds the budget (in terms of time or money).  This can be seen as a problem of inefficiency which requires acceleration of the annotation or model learning process. 
 \item[B.] The data quality does not match the requirements.  Annotators will always provide inconsistent or incorrect annotations to some extent. The process generally becomes more error-prone with the complexity of the task.  The linguistic level of the text, the number of classes to choose from, and whether the class definitions are distinct, are examples of factors that contribute to complexity.  
\end{enumerate}
Our annotation project was affected by those issues as well. In the next section, we describe a few methods that we used to mitigate them.

\section{Methods to improve the annotation process \label{sec:methods}}

Our starting point is iterative annotation in batches. In contrast to what is shown in the left panel of Fig.~\ref{fig:annotation_strategies}, the labelled dataset is hence scaled up gradually. This procedure, which we will refer to as \textit{default strategy}, allows one to train a model on intermediate data after each step and to study the development of its performance on a held-out test dataset.
It is the basis for the more advanced methods to be described in the following subsections.

\subsection{Bootstrapping \label{subsec:method_bootstrapping}}

In the default strategy, the model has a passive role and the mere purpose of training it is to track its progress in performance.  At each iteration, the annotation is done from scratch, i.e.  the annotators start from pure text data without any annotations.  Bootstrapping essentially means that one lets the model predict on a batch of data before it is sent to the annotators. This way, the annotation task shifts towards the correction of model predictions. At the beginning of the iterative process, this is not much of an advantage as the model performance is usually rather poor. However, later in the process, the model predictions are of valuable help and have the potential to significantly simplify and speed up the annotation process.  
A risk that comes with bootstrapping is that annotators may trust the model predictions too much (in later stages of the iterative process) and fail to thoroughly correct them.  This can lead to imperfect annotations and a feedback loop where the model is fed its own predictions. 

In order to make the most of bootstrapping and maximize the acceleration of the annotation process, it is imperative to train a near-optimal model after each step. This is challenging,  however, as the amount of data changes after each step, from tiny in the beginning to rather large in the end.  As the optimal number of training epochs heavily depends on the amount of data \citep{zhang2021revisiting, mosbach2021stability}, using the standard method of fine-tuning \citep{devlin2019bert, wolf-etal-2020-transformers} implies that a hyperparameter search is required to find the optimum, every time the model is trained.  To avoid the considerable effort that comes with this, we used our own fine-tuning method based on early stopping and a custom learning rate schedule \citep{stollenwerk2022adaptive}. It automatically trains for a near-optimal number of training epochs, and can be applied throughout the whole bootstrapping process. 

\subsection{Active Learning \label{subsec:method_active_learning}}

Another well-established method to accelerate the annotation process is active learning, see e.g.~\citep{settles2009active, olsson_literature_survey} for a general introduction.  The default process samples each batch randomly from the available raw data.  In contrast, active learning aims to select the batch of data that is most informative for the model. The idea is that this enables the model to learn faster, often resulting in a steeper learning curve (see e.g.~\citep{alps}). However, note that this expectation is mainly based on empirical results, and there is no guarantee that active learning helps in every single use case. In fact, there are also cases where active learning has shown to have no or even a negative effect (see e.g.~\citep{schein}).

There are many different variants of active learning. They can be categorized according to whether they use uncertainty or diversity sampling (or both),  which family of models they can be applied to (general models, neural networks, language models), and their very definition of informativeness. 
Since there were no published results on active learning in conjunction with Swedish language models when we started working on this project, we did some experiments ourselves \citep{joey}. We simulated the iterative annotation process using publicly available ground truth datasets for named entity recognition in Swedish, and tried out some of the active learning methods using different (acquisition) batch sizes.  We observed a modest positive effect of active learning \citep{joey}, and decided to use the simple query-by-uncertainty method for our own job ad annotation. Note that this course of action is based on the assumption that the effect of active learning on our own job ad dataset is similar to what we observed in our experiments with the aforementioned ground truth datasets.  However, it is important to note that there is no guarantee for this, especially as our job ad dataset differs from the ground truth datasets by having a very specific theme and a more complex class system. Moreover, there is no way to verify in hindsight that the assumption was correct without enormous additional effort\footnote{One would have to annotate a second, randomly sampled dataset under the very same conditions as the original, actively sampled dataset.}.

\subsection{Annotation Cross-checking \label{subsec:method_manual_annotation}}

As explained in Sec.~\ref{sec:introduction}, the task to annotate job ads is quite hard.  This became apparent early in our initial experiments, where we found a very low inter-annotator agreement. This made a simple annotation approach (cf.~Fig.~\ref{fig:annotation_strategies}) unsuitable for the task as it would have led to inconsistent annotations and thus low quality data.  Inevitably, some sort of cross-checking was needed to overcome the problem.  We aimed for an approach where every job ad was annotated by at least two annotators. Moreover, we wanted it to include a part where all annotators collectively annotate and have the opportunity to discuss examples together. The idea behind this is that it helps the annotators to learn how to annotate consistently, and to get feedback in order to successively improve the annotation guidelines and refine the class system (more on the latter in Sec.~\ref{subsec:method_class_system_adjustments}).  Last but not least, we aimed for a procedure that---despite involving the aforementioned elements that add complexity---was as time-efficient and simple as possible. 
We came up with the annotation strategy depicted in the right panel of Fig.~\ref{fig:annotation_strategies}.
Each batch of data is annotated in three stages:
\begin{enumerate}
 \item \textit{Individual annotation}: The batch is divided into N parts, where N is the number of annotators.  Each part is duplicated and assigned to two different annotators.  Each annotator works on their two parts individually. The result is that each part of the batch comes with two different annotations. 
 \item \textit{Merge (automated)}: Each part of the batch is "merged". To begin with, this means that its two annotations are compared. The entities that coincide are simply kept.  If the two annotations differ, for instance as a word was tagged by annotator A but not by annotator B, the word in question is turned into a special entity with the label \texttt{???}, indicating that there is disagreement that needs to be resolved.  In the case of overlapping annotations, both variants are kept and assigned the special label. After this step, there is only one (merged) annotation for each part of the batch.
 \item \textit{Collective annotation}: The annotators go through the merged parts of the batch together, discuss the cases where the annotation differed, and resolve the conflicts.  The result is considered ground truth data.
\end{enumerate}

\begin{figure}[h]
 \hspace{-2mm}
 \includegraphics[scale=0.239]{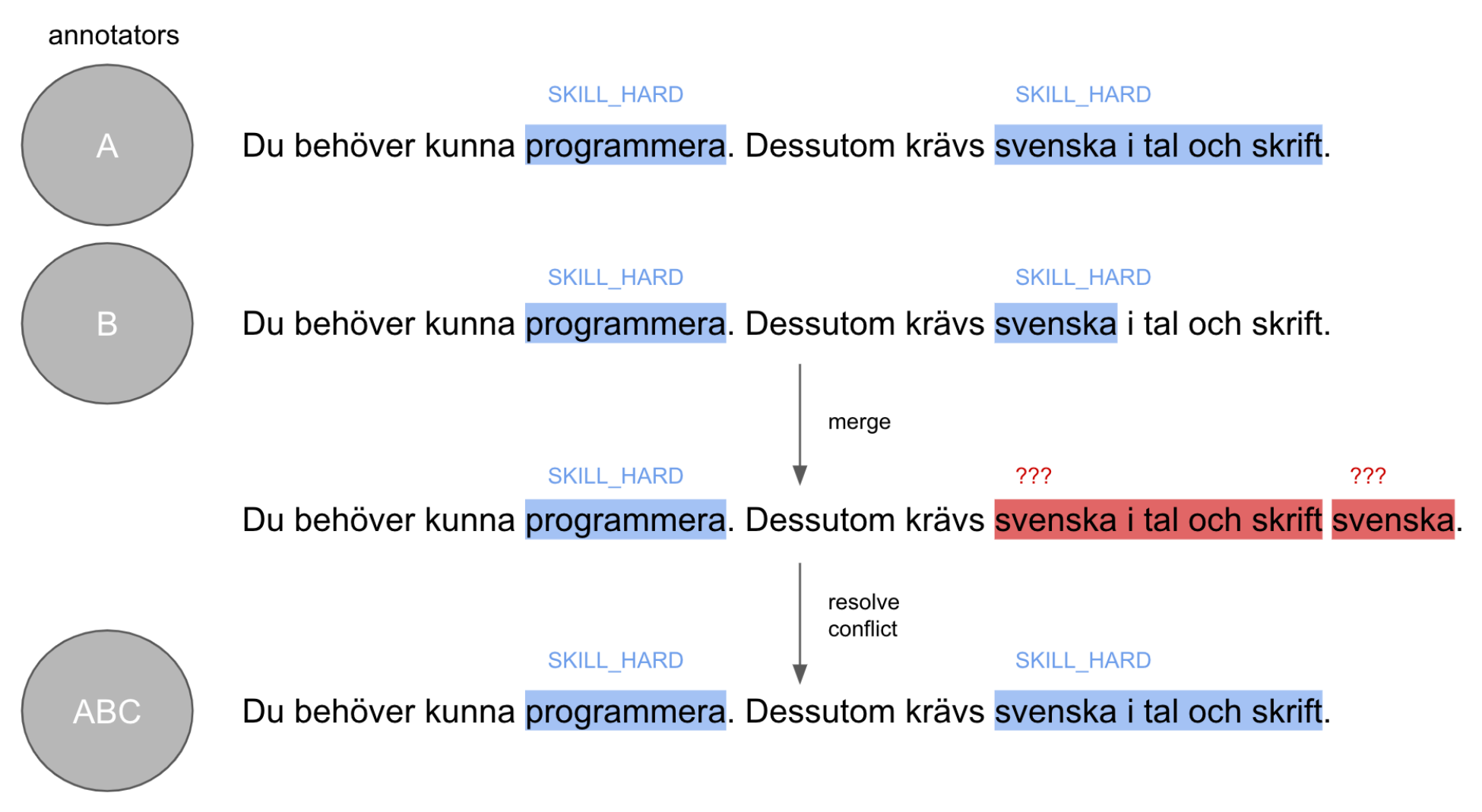} 
 \caption{Excerpt of a job ad that gets annotated separately by two annotators A and B. The result is merged before the conflicts are resolved together by all annotators. }
 \label{fig:annotation_strategy_merge}
\end{figure}

An example excerpt of a job ad annotated according to this procedure is illustrated in Fig.~\ref{fig:annotation_strategy_merge}.

During the whole process, we tracked the inter-annotator agreement using various metrics, see \citep{anna} for details.

\subsection{Class System Adjustments \label{subsec:method_class_system_adjustments}}

The initial class system that we chose before starting the annotation process was designed to match the needs of stakeholders and the downstream applications that would eventually make use of the NER model predictions.  The system was quite ambitious with 16 different classes, and it was by no means obvious to us how well the model would learn to predict them.  With this in mind, we incorporated the opportunity to dynamically adjust the class system during the iterative process in our framework. 
After each iteration, we used the following sources to evaluate how well the process worked for each class:
1. discussion after the collective annotation session (qualitative feedback)
2. inter-annotator agreement (quantitative feedback)
3. model performance, e.g.  in terms of $f_1$ score and the confusion matrix (quantitative feedback).
If we found one or more classes to be difficult and saw no signs of improvements over multiple iterations, we adjusted the class system.  Such an adjustment is defined as a mapping of the problematic classes to (presumably) unproblematic classes, including \texttt{O}.  In particular, one may simply drop a class (\texttt{A} $\to$ \texttt{O}), incorporate a class into another, existing class (\texttt{A} \& \texttt{B} $\to$ \texttt{B}) or merge two classes to form a new class (\texttt{A} \& \texttt{B} $\to$ \texttt{C}).
Note that a class system adjustment always represents an irreversible\footnote{The existing annotated data can easily be mapped from an old to a new class system.  However, if an additional batch is annotated using the new class system, it can not be mapped back to the old class system.} simplification.
Our class system was adjusted several times and we ended up with 10 instead of 16 classes. For more details and quantitative results, we again refer to \cite{anna}.

\section{Model Performance \label{sec:results}}

A total of 260 job ads were annotated by 5 people using the outlined methods. 200 of them were used to fine-tune KB-BERT, while 30 were held out as a validation dataset for hyperparameter search (see Sec.~\ref{subsec:method_bootstrapping}). The remaining 30 job ads served as a test dataset to assess the model's performance.
Following standard practice for NER, we used the $f_1$ score on the entity level as the main evaluation metric. We determined it separately for each individual class and used the micro-average as a single number to quantify the model's general performance. Both model training and evaluation were done using the \textit{nerblackbox} package \citep{nerblackbox}. The results are shown in Tab.~\ref{tab:results}.

We find a reasonable performance on most of the classes, specifically the ones that are most important for downstream applications. However, the numbers are not comparable with those typically reported on NER benchmark datasets, see e.g.~\citep{swedish-bert}. We attribute this to the increased complexity of the problem, i.e. the large number of classes with intricate, overlapping definitions (that even humans struggle with, as evidenced by the low inter-annotator agreement). 
An additional, particular challenge are classes that contain many multi-word entities, most prominently the \texttt{JOB\_TASK} class. The model often correctly identifies entities, but struggles to predict their boundaries accurately. This is reflected by the significantly better $f_1$ score on the token-level that is observed for those classes, see Tab.~\ref{tab:results}. 
\begin{table}[ht]
    \small
    \centering
    \begin{tabular}{lcc}
    \hline
        \textbf{Class} & \textbf{entity} & \textbf{token} \\ \hline
        \texttt{SKILL\_HARD} & 0.77 & 0.87 \\
        \texttt{SKILL\_SOFT} & 0.78 & 0.76 \\
        \texttt{JOB\_TITLE} & 0.90 & 0.88 \\
        \texttt{JOB\_LOCATION} & 0.76 & 0.76 \\
        \texttt{EMPLOYER\_TITLE} & 0.84 & 0.86 \\
        \texttt{JOB\_TASK} & 0.48 & 0.74 \\
        \texttt{EDUCATION\_DEGREE} & 0.62 & 0.85 \\
        \texttt{JOB\_TIME} & 0.66 & 0.85 \\
        \texttt{EXPERIENCE\_DURATION} & 0.62 & 0.80 \\
        \texttt{EMPLOYER\_BENEFIT} & 0.73 & 0.63 \\ \hline
        \textbf{micro-average} & \textbf{0.72} & \textbf{0.79}
    \end{tabular}
    \caption{Performance of our NER model. The numbers represent the $f_1$ score evaluated on the test dataset, on the entity and token level. The classes are ordered by estimated importance for downstream applications.}
    \label{tab:results}
\end{table}

\section{Conclusions \label{sec:conclusions}}

This case study highlights some typical challenges one may face when annotating real world data. We covered some advanced methods that helped us to accomplish the task. The resulting NER model shows a reasonable performance in consideration of the intricacy of the problem.    

\paragraph{Acknowledgements}

This work was done in conjunction with the Vinnova-funded research project \textit{Språkmodeller för svenska myndigheter} (Language models for Swedish authorities). 
%

\bibliographystyle{acl_natbib}
\bibliography{references}

\end{document}